\documentclass[12pt]{article}

\usepackage{graphicx}
\usepackage{amsmath}
\usepackage{amsfonts}
\usepackage{subfig}
\usepackage{float}
\usepackage{tikz}
\usepackage{fullpage}
\usepackage[hidelinks, breaklinks]{hyperref}
\usetikzlibrary{shapes.geometric, arrows, positioning}

\tikzstyle{state} = [rectangle, rounded corners, minimum width=3.5cm, minimum height=1cm, text centered, draw=black, fill=blue!20]
\tikzstyle{decision} = [diamond, minimum width=3.5cm, minimum height=1cm, text centered, draw=black, fill=orange!20]
\tikzstyle{arrow} = [thick,->,>=stealth]

\usepackage{amssymb}

\newcommand{\ie}{\textit{i.e.}, }
\newcommand{\eg}{\textit{e.g.}, }
\newcommand{\etal}{\textit{et al.}}

\newcommand{\nostarnote}[1]{}

\DeclareMathOperator*{\argmin}{arg\,min}

\title{\LARGE \bf One-Shot Gesture Recognition for Underwater Diver-To-Robot Communication}

\author{Rishikesh Joshi$^{1}$ and Junaed Sattar$^{2}$\\
\small Department of Computer Science and Engineering, Minnesota Robotics Institute,\\
\small University of Minnesota -- Twin Cities, Minneapolis, MN, USA.\\
\small {\tt joshi398@umn.edu, junaed@umn.edu}}

\date{}

\begin{document}

\maketitle

\begin{abstract}

Reliable human-robot communication is essential for effective underwater human-robot interaction (U-HRI), yet traditional methods such as acoustic signaling and predefined gesture-based models suffer from limitations in adaptability and robustness. 
In this work, we propose One-Shot Gesture Recognition (OSG), a novel method that enables real-time, pose-based, temporal gesture recognition underwater from a single demonstration, eliminating the need for extensive dataset collection or model retraining. 
OSG leverages shape-based classification techniques, including Hu moments, Zernike moments, and Fourier descriptors, to robustly recognize gestures in visually-challenging underwater environments. 
Our system achieves high accuracy on real-world underwater video data and operates efficiently on embedded hardware commonly found on autonomous underwater robots (AUVs), demonstrating its feasibility for deployment on-board robots. 
Compared to deep learning approaches, OSG is lightweight, computationally efficient, and highly adaptable, making it ideal for diver-to-robot communication. We evaluate OSG's performance on an augmented gesture dataset and real-world underwater video data, comparing its accuracy against deep learning methods.
\nostarnote{$\times$ Some reviewer will ask how the accuracy is compared to deep methods. Do we have an answer?}
Our results show OSG’s potential to enhance U-HRI by enabling the immediate deployment of user-defined gestures without the constraints of predefined gesture languages.
\end{abstract}

\section{Introduction}
\label{sec:introduction}

\nostarnote{$\times$A common rookie mistake (so it's understandable): quotes in LaTeX begin with two back quotes and end with two single quotes, like ``this'', and not like "this", the latter of which makes them appear backwards. This is the case with all of your quotes, so do fix them. Also, please do not use [H] for figure placement with the capital `H'. Really does not play nice with equations and other formatting commands, so opt for the lower-case versions instead, \ie [h].}
Research in companion Autonomous Underwater Vehicles (co-AUVs) has seen significant advances in recent times. 
These co-AUVs, being small and person-portable, make it possible to use them as ‘dive-buddies’ to humans underwater \cite{Sattar2009RSS, sattar2007where}, opening up possibilities for underwater human-robot collaboration in many different tasks, \eg environmental monitoring, species conservation, invasive species detection and removal, infrastructure inspection, and more \cite{auv_cleaning, duarte2016OCEANS}.\nostarnote{$\times$you need to cite some of these use-cases}
In all these use cases, reliable human-AUV communication methods are essential. 
However, the challenges imposed by the underwater medium cause severe signal attenuation in the electromagnetic band, rendering wireless communication unreliable and unusable beyond very close distances \cite{Birk2022, Sattar2018JFR-Islam-MotionGestures}.\nostarnote{$\times$cite these and similar claims, please} 
Acoustic signals also tend to distort, both in spoken language and tonal sounds, and are often difficult for robots to understand~\cite{Sattar2023ICRA-Fulton-HREye}. 
Gesture-based nonverbal communication from divers to robots has thus been the most used \cite{Sattar2007ICRA,chavez,Sattar2018JFR-Islam-MotionGestures}.
\nostarnote{I've been doing UHRI since long before any of these other folks, so I think its fair to cite our own work here, as this is not just a case of rampant self-citations} 
Gestural languages for underwater human-robot interaction (U-HRI) are often faced with a trade-off between expressivity and simplicity; the expressivity is necessary for divers to instruct the robots to perform a variety of tasks. 
On the other hand, an overly complicated protocol can induce communication errors, be difficult to train divers in, and add to their already high cognitive load. 
Scuba divers have a fairly high workload maintaining dive gear and life-support equipment underwater, so any method that adds minimally to this workload is generally desirable \cite{Sharma2023}. 
\nostarnote{$\times$Can you cite a reference to this workload?}

\begin{figure}[t]
	\vspace{7pt}
	\centering
	\includegraphics[width=0.5\linewidth]{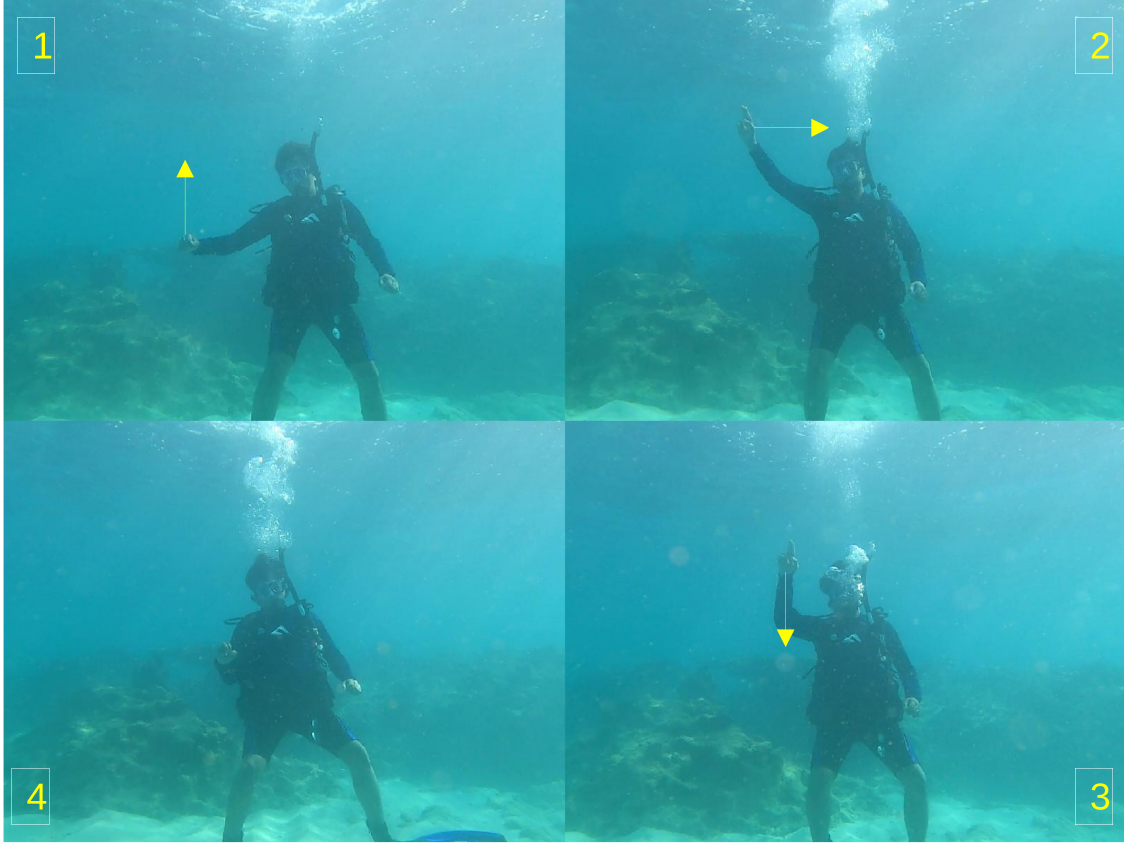}
	\caption{Clockwise starting from the top-left: a four-image sequence demonstrating the `rectangle' gesture, as defined by the diver, using the OSG approach, during open-ocean trials in the Caribbean Sea. Unlike most U-HRI gestures, OSG is spatio-temporal in nature, and uses upper body limb movements to define non-verbal cues.}
	\label{fig:rect_gesture}
	\vspace{-15pt}
\end{figure}

While gesture-based communication has shown great effectiveness, current gesture recognition systems suffer from several limitations. 
Most of the existing approaches rely on predefined gesture languages \cite{chavez}, which results in low adaptability in trained gesture recognition models. 
Introducing new gestures would require additional dataset collection (and/or synthetic data generation) and model retraining, which are cumbersome tasks. 
To address these challenges, we present One-Shot Gesture Recognition (OSG), a novel approach that enables real-time gesture recognition from a single demonstration, without the need for dataset generation or model retraining. 
OSG utilizes shape-based classification techniques to robustly recognize gestures using only their inherent structural properties, making it highly generalizable to new gestures without additional training. 
By eliminating the reliance on predefined datasets and reducing the computational burden of retraining, OSG significantly enhances the flexibility and scalability of underwater human-robot communication systems. OSG achieves $98$\% accuracy on one set of gestures and an accuracy of $89$\% on a larger gesture language in experiments on real-world underwater video data while maintaining computational efficiency, pointing to its utility as a practical, real-time solution for diver-to-robot interaction. Our contributions are:
\begin{itemize}
    \item a one-shot gesture recognition algorithm for diver-AUV interaction that requires only a single demonstration per gesture,
    \item real-world underwater video validation demonstrating high accuracy and efficient onboard execution, and
    \item a publicly available ROS2-based implementation for easy integration.
\end{itemize}

\section{Related Work}
\label{sec:related_work}

Previous human-to-robot communication methods in U-HRI relied primarily on hand signals and acoustic transmissions, which are restricted by range and environmental influences. 
As a result, researchers have been investigating ways to improve and invent new U-HRI methods--particularly gesture-based ones--to enhance human-robot interaction. 
In their U-HRI survey, Birk \etal~\cite{Birk2022} state that gesture-based systems offer a more direct and natural way of communicating with robots, even though methods such as acoustic communication, and tactile cues are commonly used. 

The CADDY project \cite{chavez} introduced a dataset that contains underwater gestures to improve U-HRI. 
The gesture language--called CADDIAN--follows a predefined set of static underwater hand gestures. 
Although this achieves the goal of gesture-based communication, it restricts divers to the CADDIAN language. 
Integrating new or custom gestures becomes challenging, as it necessitates additional data collection and model retraining. 
Moreover, static hand gestures are inherently limited by the number of unique hand configurations that the diver can comfortably perform. 
In contrast, full-arm dynamic gestures offer a larger and more expressive set of possible gestures.  

Several approaches have explored dynamic gesture-based communication in underwater environments. Sattar \etal~\cite{sattar2007where, Sattar2009RSS} analyzed vision-based control and biological motion identification for underwater human-robot interaction, which established a strong foundation for gesture recognition work. Studies like \cite{Sattar2018JFR-Islam-MotionGestures, enan2022iros} explored human-motion analysis, diver recognition, and underwater gesture-based communication. This work demonstrates the feasibility of gesture-based frameworks, especially in the underwater domain.

Ghader \etal~\cite{ghader} proposed a classification approach that uses 2D geometric features of skeleton-based images. 
Their method involved using monocular RGB input and OpenPose \cite{openpose} for diver pose keypoint detection, and a three-layer Bi-LSTM \cite{bi-lstm}\nostarnote{$\times$cite this too} neural network for gesture classification. 
The model was trained on $200$ videos, achieving a maximum accuracy of $55$\% across $11$ unique gestures in a controlled pool environment. 
However, this method relies on a pipeline that requires data collection, model training, and hyperparameter tuning. 
This, similar to the CADDY project, makes it difficult to scale beyond a predefined set of gestures. 
This is a shared limitation with dataset-driven methods.

To address this issue, Cabrera and Wachs \cite{cabrera} proposed a one-shot gesture learning approach. 
Their method generates synthetic training gesture data and classifiers such as hidden Markov models (HMM), support vector machine (SVM), and conditional random fields (CRF) to achieve accuracies of $90-93$\%. 
Although this approach is lightweight compared to deep learning models, it requires offline dataset generation and classifier training. 
Moreover, this approach requires gesture-specific parameter tuning to optimize classifier performance, as the effectiveness of the generated dataset depends on how well it represents the intended gestures.

In summary, current gesture recognition approaches are largely constrained by their reliance on either pre-trained deep learning models or extensive dataset generation and retraining. 
These are both time-consuming and computationally expensive, making it inconvenient to add custom gestures to a pre-defined gesture language. 
These challenges emphasize the need for a framework that is scalable, adaptable, and allows for real-time custom gesture integration without requiring classifier retraining or deep learning models. 
Xia and Sattar \cite{diver_recognition} demonstrate that robust underwater identification of individual divers can be achieved without relying on deep neural networks and instead using handcrafted feature descriptors and clustering-based recognition. 
This inspires OSG's approach, which similarly avoids dataset-driven training. 
Our proposed system builds upon this idea by enabling one-shot learning of gestures in real time. 
\nostarnote{Interesting that you omitted all my and the lab's prior work in underwater human-to-robot communication. any particular reason? Bilal (Ghader) is definitely not the pioneer in this line of work that your literature review makes him appear to be. The important bit is to fairly represent contributions of people in this discipline, and at this form, it does not do it.}

\section{Problem Statement}
\label{sec:problem_statement}

The objective of this project is to enable real-time recognition of user-defined gestures for U-HRI without relying on dataset augmentation or model retraining.\nostarnote{$\times$what do you mean by expansion, as opposed to collection, which you used in the intro?} 
Given a sequence of human body joint `keypoint' (\eg shoulder, wrist, hips etc.)\nostarnote{what is a keypoint here? does my addition make sense?} locations representing a gesture performed by a diver, the system must determine the most similar gesture from a given gesture language/vocabulary \nostarnote{or, `vocabulary'}.

\subsection{Input}
A \textit{gesture} can be represented as a series of recorded human body joint keypoint locations, using a suitable pose estimator (\eg YOLOv11 \cite{yolov11}, Mediapipe \cite{mediapipe}, or OpenPose \cite{openpose})\nostarnote{again, I get what you mean by keypoints but you need to specifically define it in the context of the OSG algorithm. What do these keypoints represent? I added some text, see if it makes sense}:
\begin{equation}
    \mathcal{G} = \{ p_t^i \mid i \in \mathcal{K}, t \in [0, T] \}
    \label{eq:gestures}
\end{equation}
where $p_t^i \in \mathbb{R}^2$ represents the position of keypoint $i$ at timestamp $t$, $\mathcal{K}$ is the set of all recorded keypoints (\eg left knee keypoint, right knee keypoint, etc.), and $T$ is the duration of the gesture. A \textit{language} $\mathcal{L}$ consists of a set of unique gestures, each represented as shown in Eq.~\ref{eq:gestures}.
\begin{equation}
    \mathcal{L} = \{ ( \mathcal{G}_i, y_i ) \mid i \in \mathcal{I}, y_i \in \mathcal{Y} \}
\end{equation}
where $\mathcal{I}$ is the index set of all gestures, $\mathcal{G}_i$ represents the gesture data associated with the index $i$, $\mathcal{Y}$ is the set of all unique gesture labels, and $y_i$ is the \textit{true} label of gesture $\mathcal{G}_i$.  

\subsection{Expected Output}
The system outputs the predicted label $\hat{y} \in \mathcal{Y}$ corresponding to the most likely gesture in the input language $\mathcal{L}$. This is given by:
\begin{equation}
    \hat{y} = \arg\max_{y \in \mathcal{Y}} \mathcal{S}(\mathcal{G}, \mathcal{G}_y)
\end{equation}
where $\mathcal{S}$ is a similarity function that outputs how closely the observed gesture matches a particular reference gesture (described in Sec.~\ref{sec:methodology}).

\section{Methodology}
\label{sec:methodology}
The OSG framework consists of two key components: a gesture language \textit{generator} and a gesture \textit{recognition} system.
OSG enables real-time gesture recognition without relying on any pre-trained classification models. 
This section details the methodology for defining the gesture language using the generator and subsequently performing gesture recognition.  

\subsection{Custom Gesture Language Generation}
\begin{figure}[h]
	\vspace{7pt}
    \centering
    \includegraphics[width=0.5\linewidth]{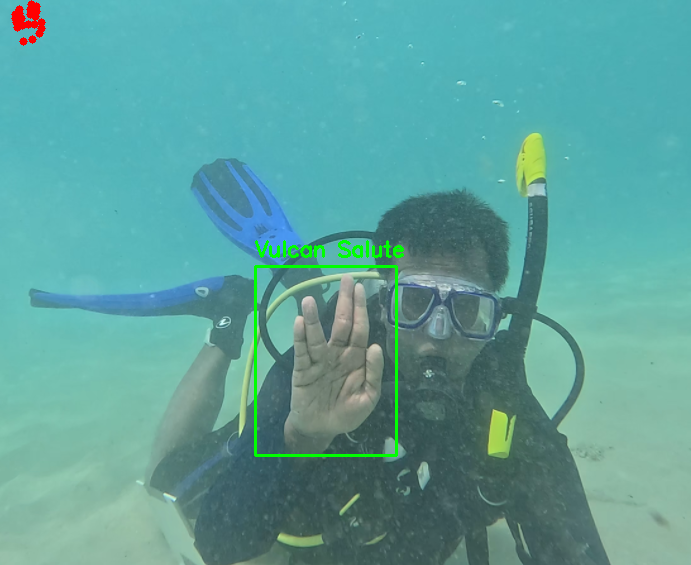} 
    \caption{The Vulcan Salute gesture, used as the start and stop trigger in OSG. The well-defined separation between the middle and ring fingers compared to the ring and pinky fingers and the index and middle fingers is what characterizes a Vulcan Salute.}
    \label{fig:vulcan_salute}
    \vspace{-10pt}
\end{figure}

The OSG framework requires a predefined yet expandable set of gestures, collectively forming a gesture language, to use as a reference for recognition. The custom gesture language is created by recording a set of gestures and extracting key features from them. A two-step process is followed in order to do this. First, the body joint keypoint locations are collected from a recorded gesture video, and second, a subset of salient joint keypoints is selected based on their movement characteristics.
The gesture recording phase begins with the detection of a start trigger, the ``Vulcan Salute'' (Fig. \ref{fig:vulcan_salute})\nostarnote{Not how you use crossreferencing in LaTeX. I edited it to show the right way.}, which activates joint keypoint tracking.\nostarnote{$\times$I think the Vulcan Salute in Barbados would make a compelling image here, instead of the one of you in the lab. Can you generate a similar bounding box for that image?} The Vulcan Salute was selected as the start trigger because it strikes a balance between ease of execution and low false positive rates in detection. Moreover, because of its uniqueness, the Vulcan Salute has a very low probability of being confused with other hand gestures divers often perform, and even with accidental hand motion, making it a robust choice for a `start-stop' trigger. Once the start trigger is detected, the keypoint locations are continuously recorded throughout gesture execution. When the gesture is completed, the Vulcan Salute is performed again to signal the end of the recording. The collected body joint keypoints, which are represented as a sequence of spatial coordinates over time, are analyzed to determine the most salient features in the gesture. \textit{Salient keypoints} are calculated using a movement-based selection process, where each keypoint's total displacement throughout the gesture is computed. Keypoints demonstrating significant movement--quantified by their peak-to-peak displacement--are selected as salient keypoints. The trajectories of these keypoints are then visualized by plotting their motion paths, which generates an image representation of the gesture. To simplify and denoise this gesture representation, the Ramer-Douglas-Peucker (RDP) algorithm \cite{rdp} algorithm is used.

This process produces a compact representation of a gesture language, where each gesture is stored as a simplified image alongside its salient keypoints. This serves as a reference for real-time gesture recognition. 
\nostarnote{$\times$Move this to the experimental section somewhere.}

\subsection{Gesture Recognition Process}
\label{gesture_recognition_process}
For real-time gesture recognition, a similar pipeline is followed:
\begin{enumerate}
    \item The system is initialized, and the predefined Vulcan Salute is executed to signal the beginning of the recording phase, during which the gesture is performed.
    \item Throughout gesture execution, body keypoints are continuously detected and tracked.
    \item Upon completion, the Vulcan Salute is performed again to mark the end of the recording phase.
    \item The recorded gesture undergoes comparison against all gestures in the predefined custom gesture language. 
    This is achieved by iterating through the gesture language and filtering the recorded keypoints based on the salient keypoints associated with each reference gesture.
    \item For each gesture in the language, the recorded keypoints are filtered according to the corresponding salient keypoints and then transformed into an image representation.
    \item The generated image is then compared to the corresponding stored image in the language using feature extraction techniques, particularly Fourier descriptors \cite{fourier_descriptors}, Hu moments \cite{hu}, and Zernike moments \cite{zernike} (detailed in \ref{sec:primary}). \nostarnote{$\times$You need to explain why you chose these, if not here, then refer to the section where you write the reasoning for these methods.}
    \item This process is repeated iteratively for all gestures in the language, after which classification is performed based on a majority voting scheme.
    \item Additional metrics such as aspect ratio, convex solidity, circularity, and path complexity are incorporated into the voting process to refine the final classification. These methods are explained in \ref{sec:secondary}.
\end{enumerate}

\begin{figure}[t]
    \vspace{7pt}
    \centering
    \resizebox{0.8\textwidth}{!}{
        \begin{tikzpicture}[
            node distance=0.8cm and 1.0cm,
            every node/.style={draw, align=center, rounded corners, minimum height=1.2cm, font=\small},
            process/.style={fill=blue!20},
            detect/.style={fill=green!20},
            feature/.style={fill=orange!20},
            classify/.style={fill=red!20},
            every path/.style={draw, thick, -latex}
        ]
            \node (start) [process] {Initialize System \\ (Vulcan Salute Start)};
            \node (detect) [right=of start, detect] {Keypoint Detection \\ and Tracking};
            \node (end) [right=of detect, process] {Gesture Completion \\ (Vulcan Salute End)};
        
            \node (features) [below=of detect, feature] {Salient Keypoint \\ Selection};
            \node (extract) [right=of features, feature] {Image Representation \\ and Feature Extraction};
            \node (classify) [right=of extract, classify] {Classification via \\ Majority Voting};
        
            \path (start) -- (detect);
            \path (detect) -- (end);
            \path (end) -- (features);
            \path (features) -- (extract);
            \path (extract) -- (classify);
        
        \end{tikzpicture}
    }
    \caption{Gesture Recognition Pipeline Visualized: The process begins with initialization and keypoint tracking, followed by gesture completion detection, feature extraction, and classification.}
    \label{fig:methodology_flowchart}
    \vspace{-10pt}
\end{figure}
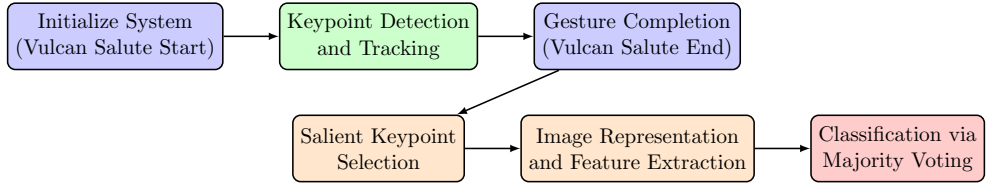

Fig.~\ref{fig:methodology_flowchart} depicts a concise visual representation of the gesture recognition process. \nostarnote{$\times$Figures need to appear on the same page, or worst case, the following page, from where you refer to them for the first time.}\nostarnote{$\times$Figures 3 and 4 are out of order.}

\subsection{Image Preprocessing}
\begin{figure}[h]
    \centering
    \includegraphics[width=0.5\linewidth]{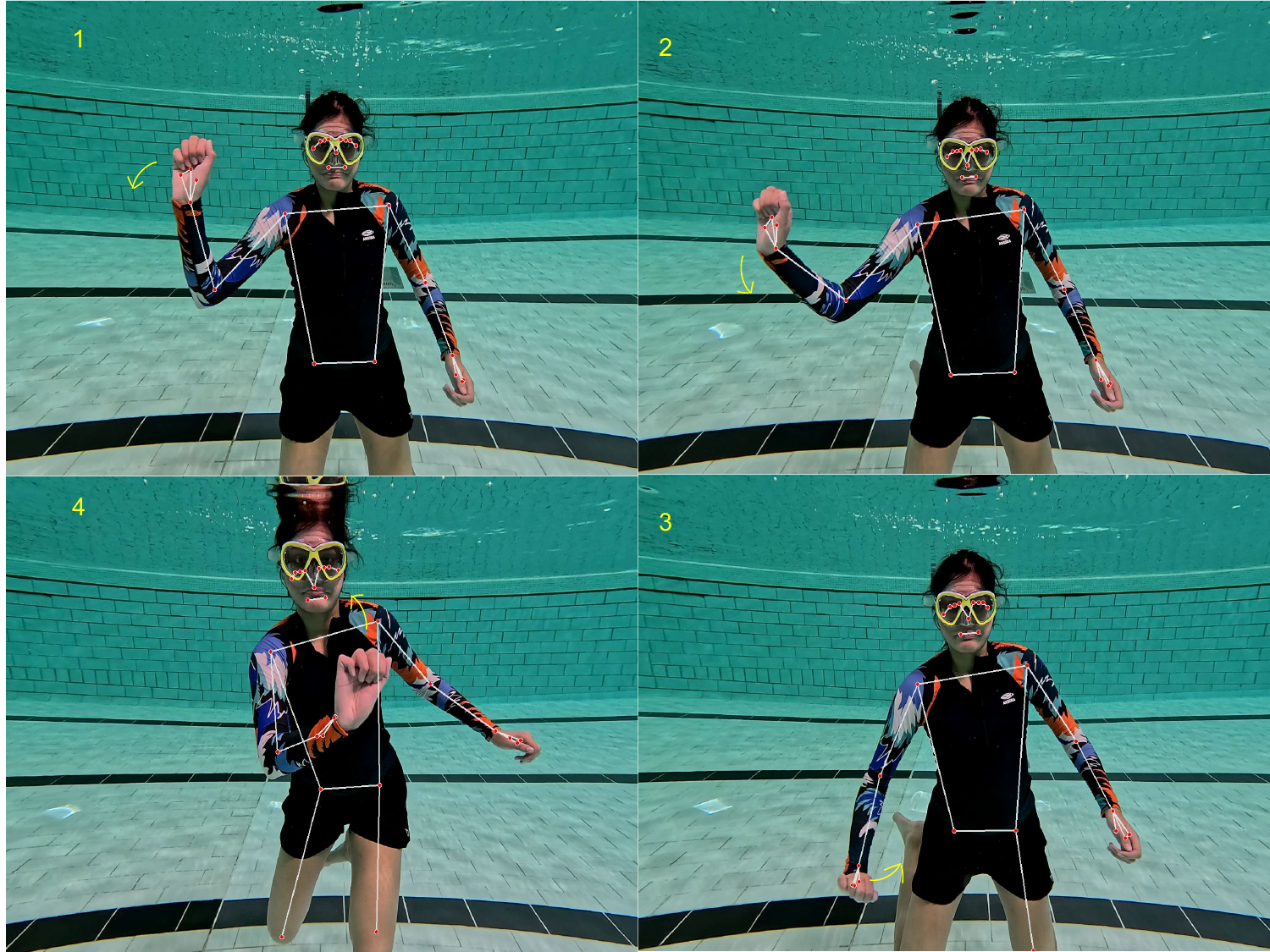} 
    \caption{Clockwise starting from the top-left: a four-image sequence demonstrating the `circle' gesture, as defined by the diver, using the OSG approach. This demonstrates the body joint keypoint tracking that OSG utilizes.}
    \label{fig:circle_gest_pool}
    \vspace{-10pt}
\end{figure}

Before feature extraction, the recorded gesture undergoes a series of preprocessing steps to normalize it and make it invariant to changes in user pose and motion.
First, all the keypoints are normalized with respect to the center of the diver's torso to account for positional changes due to movement caused by ocean currents or natural body adjustments during the dive.\nostarnote{$\times$Why pressure? Diver's don't deform, and if they do, they're dead.} 
This step renders gestures invariant to global translation, meaning that changes in the diver's position within the scene--either due to ocean currents or robot/camera movement--do not affect recognition.\nostarnote{$\times$``Global'' as in? Camera/robot movement? If so, we should say that to avoid confusion and ambiguity.}
Once this is done, keypoints are recentered based on their centroid to make sure that they are aligned with respect to a fixed reference position.\nostarnote{Seeing how important the keypoints are in the algorithmic description, I think you need to show visuals of the pose and keypoint detection in a figure somewhere near this text, for better understanding.}
Fig.~\ref{fig:circle_gest_pool} shows an example of the human pose estimation and associated keypoints during a `circle' gesture. 
The keypoints are then plotted onto a graph to generate an image representation of the gesture. Additionally, the RDP algorithm \cite{rdp} is applied to denoise the gesture image while preserving its key shape characteristics. 

\subsection{Voting-Based Classification:}
\label{sec:voting}
OSG evaluates the similarity of the recorded gesture by computing the distances between its image moments and Fourier descriptors to those of the stored language gestures, which we refer to as reference gestures. Each method selects the reference gesture with which it has the lowest computed distance, casting a vote for that gesture. The final classification is determined by the majority vote among these methods.

\subsubsection{Primary Voters} 
\label{sec:primary}
The primary voters in OSG extract features that encode the entire gesture representation into descriptors. This allows for direct comparison using Euclidean distances. While cosine similarity can be used as a comparison metric, it disregards magnitude differences in feature vectors, which are crucial in distinguishing gestures. Since our method relies on shape descriptors that encode both structure and size, Euclidean distance is preferred as it captures absolute differences in gesture execution. \nostarnote{$\times$Would something like the Cosine similarity be more robust? Asking for a reviewer.}These methods include:

\begin{enumerate}
    \item \textbf{Hu Moments:} Hu moments are a set of 7 invariant descriptors that encode shape properties while providing robustness against translation, rotation, and scaling \cite{hu}. The central moments of an image $I(x,y)$ are given by:
    \begin{equation} 
        \mu_{pq} = \sum_{x} \sum_{y} (x - \bar{x})^p (y - \bar{y})^q I(x, y) 
    \end{equation} 
    where $(\bar{x}, \bar{y})$ is the centroid of the shape. Hu's 7 invariant moments are calculated by applying various nonlinear transformations to these central moments. This method enables OSG to compare gestures based on their overall \textit{structure}. 
    \item \textbf{Zernike Moments:} Zernike moments use an orthogonal basis set of polynomials that capture intricate shape characteristics with higher precision \cite{zernike}.
    \begin{equation}
        V_n^m(\rho, \theta) = R_n^m(\rho) e^{jm\theta}
    \end{equation}
    where $n$ is a non-negative integer, $m$ is an integer such that $n - |m|$ is even, $\rho \in [0, 1]$ is the normalized radial coordinate, and $R_n^m(\rho)$ is the radial polynomial\nostarnote{$\times$equation exceeds margin below}:
    \begin{align}
        R_n^m(\rho) &= \sum_{s=0}^{(n - |m|)/2} 
        \frac{(-1)^s (n-s)!}
        {s! \left( \frac{n+|m|}{2} - s \right)! \left( \frac{n-|m|}{2} - s \right)!}\rho^{n-2s}
    \end{align}
    Zernike moments are then computed as:
    \begin{equation}
        Z_n^m = \frac{n+1}{\pi} \int_{0}^{2\pi}\int_{0}^{1} I(\rho, \theta) V_n^m(\rho, \theta)^* \, \rho \, d\rho \, d\theta.
    \end{equation}
    where \(V_n^m(\rho, \theta)^*\) is the complex conjugate of the Zernike polynomial.
    These moments provide an informative representation of \textit{shape} features.
    \item \textbf{Fourier Descriptors:} Fourier descriptors encode the contour of gesture trajectories in the frequency domain \cite{fourier_descriptors}. This method is effective in recognizing shapes that involve smooth and continuous trajectories.
    \begin{equation}
        z_k = x_k + j y_k, \quad k = 0,1,\dots,N-1.
    \end{equation}
    Taking the Discrete Fourier Transform (DFT) of this sequence gives:
    \begin{equation}
        Z_u = \sum_{k=0}^{N-1} z_k e^{-j 2\pi uk/N}, \quad u = 0,1,\dots,N-1.
    \end{equation}
    The Fourier descriptors are then obtained by selecting the low-frequency coefficients. These coefficients are particularly selected because they summarize the key shape characteristics of the gesture while filtering out high-frequency noise and minor gesture execution variations.\nostarnote{$\times$say a word about why you just need the LF coefficients.}
\end{enumerate}

\subsubsection{Secondary Voters}
\label{sec:secondary}
To improve the accuracy of OSG's classification, additional features are considered in the voting process. These methods do not generate vector encodings of images. Instead, they function as low-level, crude quantifications of gesture shapes:
\begin{enumerate}
    \item \textbf{Aspect Ratio:} The ratio of the width and height of the bounding box enclosing the gesture image. This is used to differentiate between gestures with elongated characteristics versus gestures that are relatively more compact.
    \item \textbf{Convex Solidity:} This measures how convex or concave a shape is. It is calculated using the ratio of the area of the contour of the gesture's shape and the area of its convex hull. This method is particularly useful for differentiating between open structures (\eg V-shape, C-shape) and solid shapes (\eg circle, triangle).
    \item \textbf{Circularity:} This is used to differentiate between circular and angular shapes. It is calculated using \( \frac{4\pi a}{p^2} \) where $a$ is the area of the shape and $p$ is its perimeter.
    \item \textbf{Path Complexity:} This measures the intricacy of the gesture's shape by comparing the total length of its contour to the perimeter of its bounding box. A higher value indicates a shape with more complex strokes (\eg zig-zag, spiral) whereas a lower value corresponds to simpler contours like circles or rectangles. It is calculated as \( \frac{\text{Total Path Length}}{\text{Bounding Perimeter}} \)
\end{enumerate}
Each of these secondary features is computed for both the recorded and reference gestures and then compared using the difference in values. 
For a given recorded gesture \(\mathcal{G}_r\), each method independently evaluates the difference $\mathcal{G}_r - \mathcal{G}_i$ with all gestures \(\mathcal{G}_i, i \in \mathcal{I}\) with \(\mathcal{I}\) being the vocabulary size. A vote is cast for the most similar gesture \(\mathcal{G}_k\) based on the smallest difference, \ie: 
\begin{equation}
    \argmin_{k\in \mathcal{I}}(\mathcal{G}_r - \mathcal{G}_k)
\end{equation}
This process is repeated for all features, treating each as an independent voter. \nostarnote{Thanks for the mattermost clarification -- see if this makes sense and edit/correct as you see fit, please.}

\begin{figure}[h]
    \vspace{7pt}
    \centering
    \includegraphics[width=0.5\columnwidth]{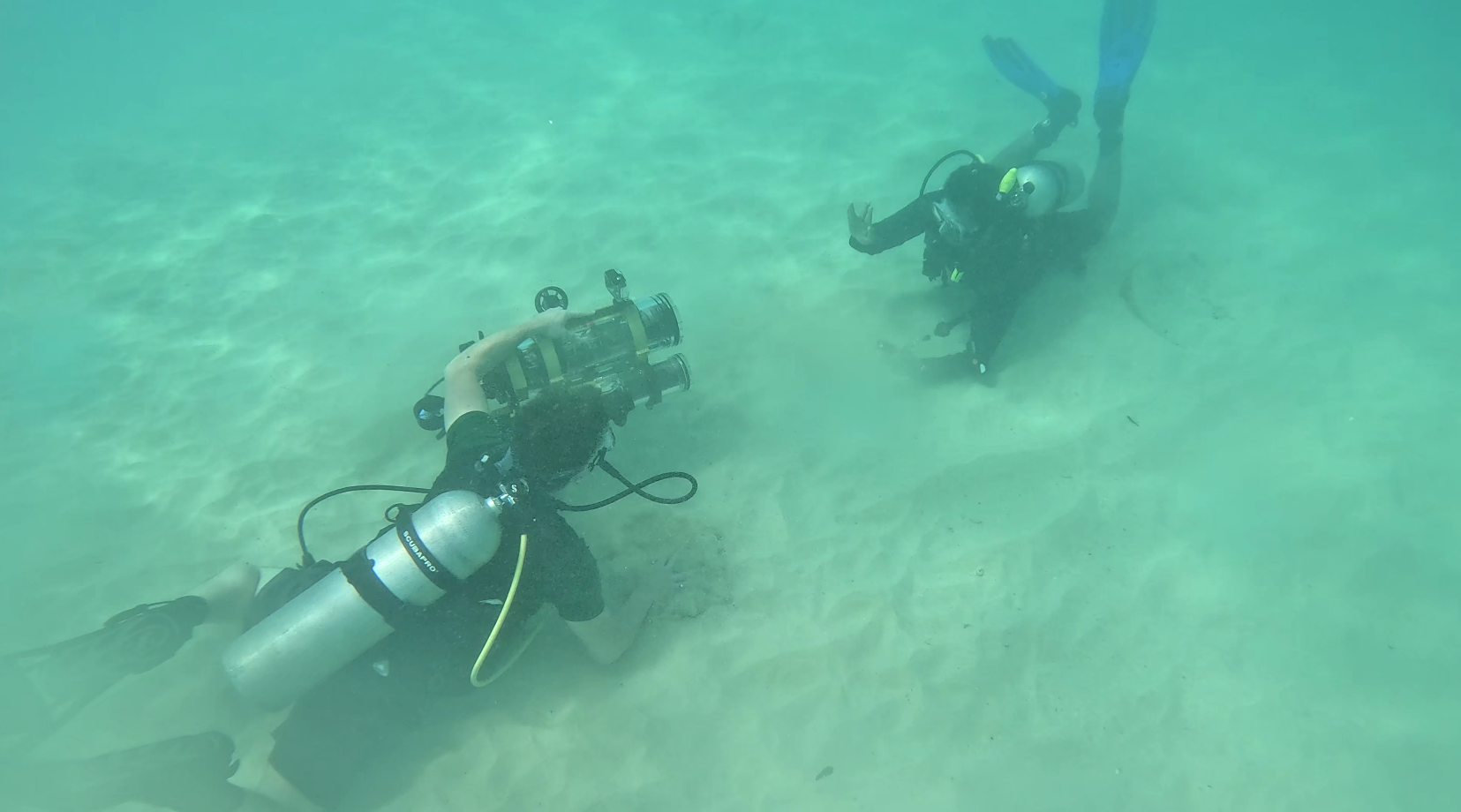} 
    \caption{A third-person view of OSG being tested during an open-ocean trial in the Carribean Sea. The image depicts a Vulcan Salute, the predefined start and stop trigger for OSG, being performed in front of the MeCO AUV \cite{MeCOProject}}
    \label{fig:confusion_matrix1.png}
\end{figure}

\section{Experiments}
\label{sec:experiments}

We evaluate the performance of OSG using video data gathered from the MeCO AUV~\cite{MeCOProject}\nostarnote{Can cite the GitHub repo, or the arXiv version when that is uploaded} in open ocean and controlled pool settings. Some of the gestures are demonstrated in the attached video, along with OSG's recognition capabilities.
Given that OSG is a training-independent framework, we measure its robustness, computational efficiency, and accuracy in classification. 
Our approach does not require any gesture datasets to train on and performs recognition based on inherent shape characteristics of gestures.

Since the number of real-world underwater gesture samples was limited due to a lack of readily available datasets and the significant effort required to create an underwater dynamic gesture dataset from scratch\nostarnote{$\times$give a reason as to why}, we used data augmentation to generate a dataset consisting of $200$\nostarnote{See how to write numbers in LaTeX in the code here.} examples of $3$ unique gestures, for a total of $600$ data points to test OSG. 
This data augmentation process followed the approach of \cite{cabrera} and preserved the fundamental motion characteristics of each gesture in our defined language (Table \ref{tab:gesture_table} and Fig. \ref{fig:gesture_language})\nostarnote{$\times$Fix the references like I demonstrated in the previous section} while also increasing the diversity in the dataset. 
The system achieved an overall accuracy of $98$\%, which demonstrates its ability to detect gestures in a reliable way. 
The results presented in the confusion matrix (Fig. \ref{fig:confusion_matrix1.png}) show the robust gesture recognition capabilities of OSG, despite natural variations in gestures.
\begin{figure}[h]
    \vspace{7pt}
    \centering
    \includegraphics[width=0.5\columnwidth]{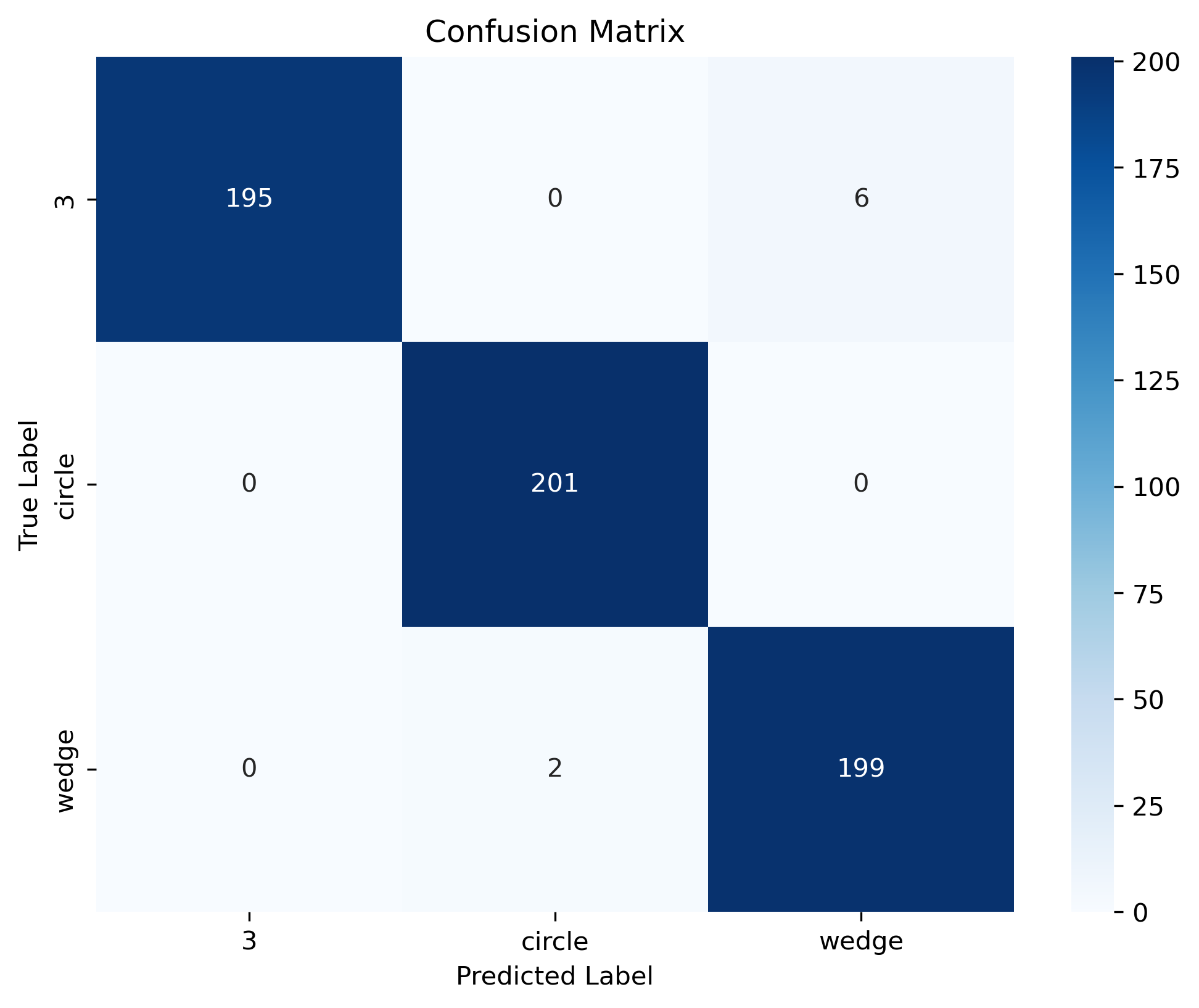} 
    \caption{Confusion matrix for gesture language 1. Each row represents the true gesture label, and each column corresponds to the predicted label. Darker diagonal elements indicate correct classifications whereas darker non-diagonal elements highlight incorrect classifications. The high accuracy can be attributed to the distinct shape characteristics of each gesture.}
    \label{fig:confusion_matrix1.png}
\end{figure}
OSG was also tested in a terrestrial setting\nostarnote{for my clarity, what does `manual testing' entail?} on a larger gesture language (Table \ref{tab:gesture_table2}). Each gesture was tested $50$ times, resulting in a total of $400$ tests. The system achieved an accuracy of $89$\% on this larger gesture set. A breakdown of the results is provided in the corresponding confusion matrix (Fig. \ref{fig:confusion_matrix2.png}). Keypoint data was collected using Mediapipe \cite{mediapipe} and Vulcan Salutes were identified using YOLOv7 \cite{yolov7}. The gestures in our custom-defined languages (Tables \ref{tab:gesture_table} and \ref{tab:gesture_table2}) were selected to include a diverse range of shape characteristics that ensured variability in circularity, angularity, and general motion complexity. For example, the `Circle' gesture represents a smooth, closed shape while the `Wedge' gesture introduces an open shape that has a sharp angle. This selection was intentionally done to test OSG's capabilities in differentiating various motion profiles and to ensure distinctiveness among gestures. 
This shows that empirically, OSG has a high success rate even for gesture languages with larger vocabularies. However, it is heavily dependent on body joint keypoint tracking, which can cause problems in underwater conditions with poor visibility. This is further detailed in \ref{sec:limitations}. 
\nostarnote{$\times$Please keep in mind underwater visibility will be a factor and you need to make the reviewers realize that you have considered such factors. Also, $98$\% is pretty high accuracy, so the reviewers will come at you hard -- need to be careful about conditions and claims on the results.} 
To measure the real-time efficiency of the system, we measured the average frames processed per second (FPS), and the power consumption. 
\nostarnote{$\times$Do I see some inconsistent boundary lines around the gesture images in Fig.~\ref{fig:gesture_language}? Can you remove them if possible?}

\begin{table*}[t]
    \vspace{7pt}
    \centering
    \renewcommand{\arraystretch}{1.2} 
    \resizebox{0.8\textwidth}{!}{
        \begin{tabular}{|c|c|c|}
            \hline
            \textbf{Gesture Name} & \textbf{Salient Keypoint(s)} & \textbf{Brief Description} \\  
            \hline
            Circle       & Right wrist       & Circular motion performed by the right hand \\  
            Wedge        & Right wrist       & V-shaped gesture using the right hand  \\  
            Three       & Right wrist      & Right hand gesture that mimics drawing a number `3' shape \\
            \hline
        \end{tabular}
    }
    \caption{A selection of gestures from a predefined gesture language used in the OSG system. Each gesture is associated with a specific salient keypoint, which is tracked for recognition. A brief description of each gesture’s motion, such as the `Circle' gesture involving a circular movement of the right wrist, and the `Wedge' gesture forming a V-shape, is provided.}
    \label{tab:gesture_table}
\end{table*}

\begin{figure}[t]
    \centering
    \resizebox{0.8\textwidth}{!}{
        \subfloat[Circle gesture]{
            \includegraphics[width=0.3\linewidth]{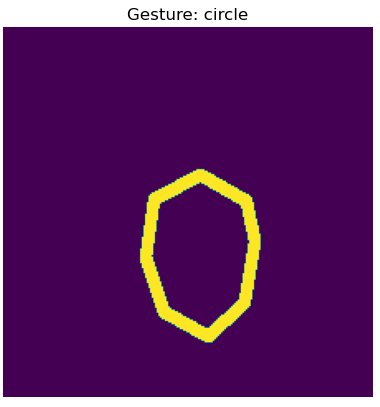}
            \label{fig:subfig1}
        }
        \subfloat[Wedge gesture]{
            \includegraphics[width=0.3\linewidth]{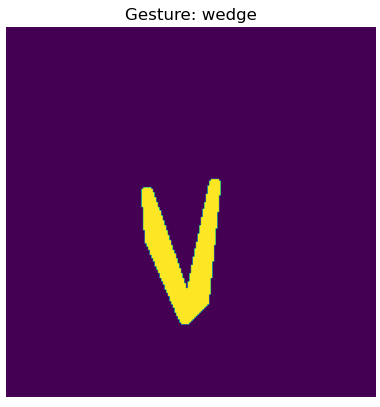}
            \label{fig:subfig2}
        }
        \subfloat[Three gesture]{
            \includegraphics[width=0.3\linewidth]{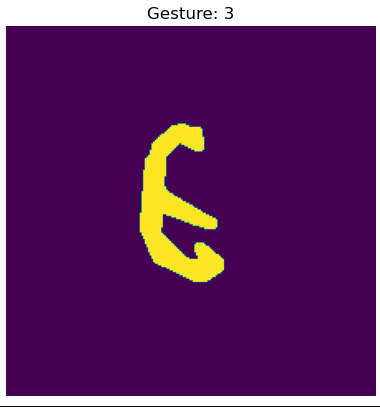}
            \label{fig:subfig3}
        }
    }
    \caption{Image representations of a custom defined gesture language. Each subfigure depicts the image representation of a gesture in the language corresponding to Table \ref{tab:gesture_table}. (a) is a circular motion of the right wrist for the `Circle' gesture, (b) is a V-shaped movement for the `Wedge' gesture, and (c) is a right-hand motion resembling the number ``3'' for the `Three' gesture.}
    \label{fig:gesture_language}
    \vspace{-10pt}
\end{figure}

\begin{figure}[t]
    \vspace{7pt}
    \centering
    \includegraphics[width=0.5\columnwidth]{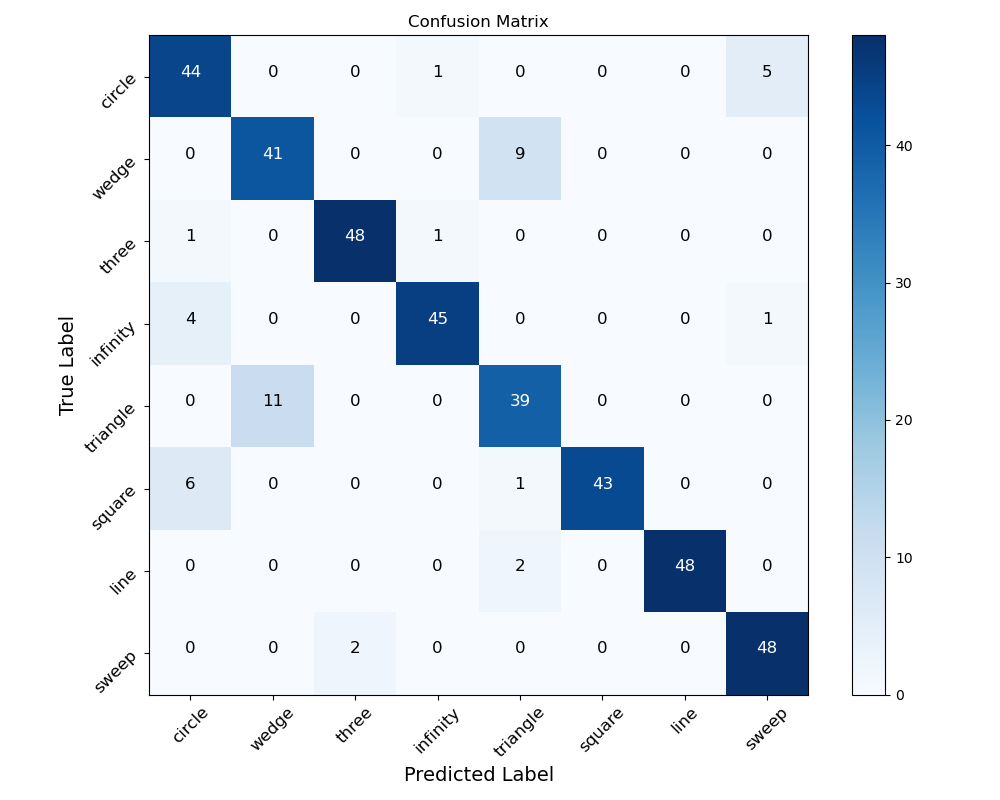} 
    \caption{Confusion matrix for gesture language 2. Each row represents the true gesture label, and each column corresponds to the predicted label. The `Triangle' and `Wedge' gestures exhibit lower accuracy rates due to their similarity in shape.}
    \label{fig:confusion_matrix2.png}
    \vspace{-10pt}
\end{figure}

\begin{table*}[t]
    \vspace{7pt}
    \centering
    \renewcommand{\arraystretch}{1.2} 
    \resizebox{0.9\textwidth}{!}{
        \begin{tabular}{|c|c|c|c|}
            \hline
            \textbf{Gesture Name} & \textbf{Salient Keypoint(s)} & \textbf{Brief Description} & \textbf{Example Usage} \\  
            \hline
            Circle       & Right wrist       & Circular motion performed by the right hand & Signaling readiness \\  
            Wedge        & Right wrist       & V-shaped gesture using the right hand & Indicating a directional command  \\  
            Three       & Right wrist      & Right hand gesture that looks like the number 3 & Signaling the number three for communication \\
            Infinity & Right wrist & $\infty$-shaped gesture & Representing continuous or looping commands \\
            Sweep & Left wrist, Right wrist & Sweep both hands up and down in an arc & Attracting attention \\
            Triangle & Right wrist & Triangle-shaped gesture & For navigation or waypoint indication \\
            Square & Right wrist & Square-shaped gesture & Alignment command \\
            Line & Right wrist & Horizontal sweeping motion & Signifying a stop to a process or plan \\
            \hline
        \end{tabular}
    }
    \caption{Outline of the larger set of gestures on which tests were conducted. This table includes the salient keypoints, used for each gesture, a brief description of its motion, and example use cases. The gestures range from simple motions like the `Circle’ for signaling readiness to relatively more complex movements such as the `Infinity’ gesture representing continuous motion.}
    \label{tab:gesture_table2}
    \vspace{-10pt}
\end{table*}

\section{Discussion}
\label{sec:discussion}
\nostarnote{Leaving this comment here for lack of a good spot. Your results images need cleanup. The pipeline figure is pretty bad IMHO. This is too cluttered, has too much text, and has too much whitespace. Also, image captions need to be descriptive. See how I captioned Figure 1. What do you want the users to focus on in your figures? Mention that in the caption. Right now, the captions are the most obvious ones. I already know Figure 3 is the pipeline, but what is it that you want the readers to focus on? Same goes for table captions too. Also, if you can, cross reference the section of the text that the figure is talking about from the caption. \eg ``General gesture recognition pipeline as described in Section BLAH''. Hope this makes sense. Also, you need more figures. Like, you show the imae representations, but not the actual gestures anywhere like I did in Figure 1. At least one would be good to have, and also, mention at the appropriate location(s) in text that you will show sequences in the attached video.}
\subsection{Robot Implementation and Deployment:}
OSG is implemented in ROS2 \cite{ros2} for robot deployment. 
The gesture recognition pipeline uses Mediapipe \cite{mediapipe} for pose estimation and YOLO \cite{yolov7} for hand detection, both running on an NVIDIA Jetson Orin edge device. 
To enhance performance, ONNX-based acceleration was used for hand keypoint extraction and tracking. 
Apart from this, the hardware requirements are minimal--only a functional monocular camera operating at approximately $15$ FPS is sufficient for real-time gesture recognition. 

Although OSG was successfully tested on the MeCO AUV \cite{MeCOProject} in a bench setup, in-water testing was hindered by external factors unrelated to OSG itself. This remains the biggest limitation of this study, as real underwater testing on-board a co-AUV would show stronger evidence of OSG's capabilities. 

All implementation details, including the ROS2 code, is publicly available here: \url{https://github.com/IRVLab/One-Shot-Gesture-Recognition} \nostarnote{mention repo here}.

\subsection{Accuracy and Comparison}
OSG achieves an accuracy of $98$\% on the first gesture language (Table \ref{tab:gesture_table}) and $89$\% on the larger gesture language (Table \ref{tab:gesture_table2}), which is on par with state-of-the-art (SOTA) machine learning models in one-shot gesture recognition that operate on a fixed gesture language. 
This is a significant result, especially considering that OSG requires only one demonstration of gesture and no data augmentation. 
It performs best when gestures have distinct shapes that are separable. 
However, highly similar gestures that only differ in execution speed present a challenge for OSG's recognition capabilities. 
This is attributed to the fact that OSG relies on shape analysis rather than temporal characteristics. 
Despite this, because OSG does not rely on labeled datasets or GPU-based inference, it serves as a compelling alternative to aforementioned gesture recognition methods in resource-scarce environments. Additionally, machine learning and deep learning pipelines often suffer from overfitting and lack interpretability--both of which OSG inherently avoids.

\subsection{Performance Analysis}
The low computational cost of OSG is one of its strengths. 
It allows for quick deployment of new gestures, without requiring re-training or data augmentation.
\nostarnote{$\times$expansion vs augmentation -- I would stick to one, and augmentation is more commonplace in literature.} 
Defining a new gesture language using the provided Python script and integrating the generated JSON file into the ROS2-based implementation is sufficient for recognizing new sets of gestures instantly. 
OSG was tested on an NVIDIA Jetson Orin and the FPS performance results demonstrate its feasibility for real-time deployment on robots. 
The system achieves a base $15$ FPS when processing only the raw camera feed, $4.91$ FPS when running keypoint detection models (pose estimation), and $4.06$ FPS when executing the full OSG pipeline, including gesture classification. 
This indicates that keypoint extraction contributes the most to computational overhead, reducing FPS by $67$\%, while our algorithm introduces an additional 17\% reduction. 
These measurements were recorded on an 8-core ARM Cortex-A78AE CPU running at $1.49$ GHz.

OSG operates well within the power constraints of an AUV. 
The average total measured power drawn from the system was approximately $8.5-9$ W, which was measured using the tegrastats utility on the NVIDIA Jetson Orin onboard the AUV \cite{tegrastats}. 
\nostarnote{$\times$Can you cite tegrastats?}This suggests that OSG is well-suited for energy-efficient, autonomous deployment. 

\subsection{Usability}
\nostarnote{$\times$I feel like this sentence can either be more specific, or omitted entirely.} 
The entire process from generating a custom gesture language to deploying it on robots follows a well-defined and intuitive pipeline that minimizes complexity in implementation.\nostarnote{$\times$What do you mean by `streamlined'?} 
Gestures are recorded, stored in JSON files alongside relevant information, and immediately deployed without requiring further modifications. 
This ease of use and customizability makes it particularly useful for HRI and U-HRI scenarios where predefined gesture languages may not be suitable for all use cases. Given that our focus is on the algorithmic design of OSG, we have not conducted human studies to rigorously evaluate its usability. This remains an open avenue for future work.
\nostarnote{$\times$need to say something about putting actual IRB-approved human-studies for usability as future work, either here or in the next section.}

\subsection{Limitations and Potential Future Work}
\label{sec:limitations}
Since OSG is purely a shape-based classifier, it struggles with gestures that vary based on speed of execution and gestures that are very similar in shape. 
It is possible to overcome this limitation with more advanced approaches, \eg with transformer-based models \cite{Montazerin2023} \nostarnote{$\times$cite one transformer-based approach you think is appropriate here}, but their applicability to underwater gesture recognition is yet unknown. Such models rely on high-density surface EMG signals, which are not available in underwater environments, making their direct application infeasible.\nostarnote{$\times$applicability in what way?}

Another limitation is OSG's lack of directionality separation in gestures. 
It does not inherently recognize trajectory-based variations in gestures. 
Future improvements could combine current shape-based classification with temporal methods employed in \cite{cabrera} to improve OSG's directional awareness while maintaining its training-free nature. Moreover, OSG is heavily reliant on keypoint detection models to perform accurate classification. This dependence poses a significant challenge in underwater environments with murky water or low visibility, where off-the-shelf models tend to fail. Custom training a YOLO-based keypoint detection model on underwater data could improve robustness \cite{Sattar2018JFR-Islam-MotionGestures, islam2018dynamic}. 

Despite these limitations, our solution fundamentally differs from modern gesture recognition frameworks, which makes direct comparisons challenging. Most SOTA methods rely access to clean datasets and on dataset augmentation, whereas OSG is designed to operate without any pre-trained classification models. While this makes it inherently more adaptable and resource-efficient for real-time applications, this distinction in approach prevents a like-for-like comparison. 
\nostarnote{$\times$You state that your approach is scalable and efficient, but you do not compare with other methods. Reviewers will give you some serious crap for that. I recommend you either say this differently, or, give clear reasons why you cannot do apples to apples comparisons with other SOTA methods.}

\section{Conclusion}
\label{sec:conclusion}

We present a real-time one-shot dynamic gesture recognition framework (OSG), designed with a focus on shape-based classification rather than large data-driven machine learning approaches. Contrary to deep learning-based approaches that require labeled datasets to be collected or artificially generated, OSG uses a deterministic recognition approach solely relying on the geometric and structural features of gestures. 
With high accuracy in\nostarnote{$\times$Avoid such self-glorifying phrases. Say ``very high accuracy'' or something more objective.} 
empirical tests, OSG is comparable to SOTA ML-based models that operate on fixed gesture languages. 
This result reinforces the idea that complex AI pipelines are not always necessary for robust gesture recognition. Additionally, OSG is a lightweight and low computational cost framework that is easy to deploy on robots and embedded systems.
The ability to define and update gesture languages instantly without retraining a model adds to OSG's usability for real-world applications. 
Although the system excels at recognizing distinct gesture shapes, its reliance on shape-based characteristics means it struggles with gestures that differ in speed or directionality. 
Beyond underwater applications, OSG can be generally useful in constrained environments where gestures serve as a primary communication method.
Future work will investigate refining OSG's classification pipeline to include sensitivity to temporal aspects\nostarnote{Why is citation [19] in all caps?}. 

\nostarnote{$\times$For a paper, this is unnecessary. You already mentioned this a few times, and we do not need a sales pitch. The results have spoken for themselves already. I would drop this paragraph. The Deep learning cool-aid drinkers will come at you with pitchforks, and that's most of the community these days.}

\bibliographystyle{plain}
\bibliography{citations}

\end{document}